\documentclass[10pt,twocolumn,letterpaper]{article}

\usepackage{iccv}
\usepackage{times}
\usepackage{epsfig}
\usepackage{graphicx}
\usepackage{amsmath}
\usepackage{amssymb}
\usepackage{subfigure}
\usepackage{multirow}
\usepackage[breaklinks=true,bookmarks=false]{hyperref}

\iccvfinalcopy 

\def\iccvPaperID{****} 

\ificcvfinal\fi

\begin{document}

\title{Masked Face Recognition Challenge: The InsightFace Track Report}

\author{Jiankang Deng\\
Imperial College London\\
{\tt\small j.deng16@imperial.ac.uk}
\and
Jia Guo\\
InsightFace\\
{\tt\small guojia@gmail.com}
\and
Xiang An\\
InsightFace\\
{\tt\small anxiangsir@gmail.com}
\and
Zheng Zhu\\
Tsinghua University\\
{\tt\small zhengzhu@ieee.org}
\and
Stefanos Zafeiriou\\
Imperial College London\\
{\tt\small s.zafeiriou@imperial.ac.uk}
}

\maketitle
\ificcvfinal\thispagestyle{empty}\fi

\begin{abstract}
During the COVID-19 coronavirus epidemic, almost everyone wears a facial mask, which poses a huge challenge to deep face recognition. In this workshop, we organize Masked Face Recognition (MFR) challenge \footnote{\url{https://ibug.doc.ic.ac.uk/resources/masked-face-recognition-challenge-workshop-iccv-21/}} and focus on bench-marking deep face recognition methods under the existence of facial masks. In the MFR challenge, there are two main tracks: the InsightFace track and the WebFace260M track \cite{zhu2021mfrwebface}.
For the InsightFace track, we manually collect a large-scale masked face test set with 7K identities. In addition, we also collect a children test set including 14K identities and a multi-racial test set containing 242K identities. By using these three test sets, we build up an online model testing system, which can give a comprehensive evaluation of face recognition models. To avoid data privacy problems, no test image is released to the public. As the challenge is still under-going, we will keep on updating the top-ranked solutions as well as this report on the arxiv.
\end{abstract}

\section{Introduction}

Recently, great progress has been achieved in face recognition with large-scale training data~\cite{guo2016ms,nech2017level,an2020particalfc,zhu2021webface260m}, sophisticated  network structures~\cite{schroff2015facenet,he2016deep} and advanced loss designs~\cite{sun2014deep,taigman2014deepface,schroff2015facenet,parkhi2015deep,deng2017marginal,liu2017sphereface,tencent2017CosineFace,wang2018additive,deng2019arcface,deng2020sub,huang2020curricularface,deng2021variational}. However, existing face recognition systems are presented with mostly non-occluded faces, which include primary facial features such as the eyes, nose, and mouth. During the COVID-19 coronavirus epidemic, almost everyone wears a facial mask, which poses a huge challenge to existing face recognition systems. Traditional face recognition systems may not effectively recognize the masked faces, but removing the mask for authentication will increase the risk of virus infection.

To cope with the above-mentioned challenging scenarios arising from wearing facial masks, it is crucial to improve the existing face recognition approaches\footnote{\url{https://pages.nist.gov/frvt/html/frvt_facemask.html}}. Generally, there are two kinds of methods to overcome masked face recognition: (1) recovering unmasked faces for feature extraction and (2) producing direct occlusion-robust face feature embedding from masked face images.

Based on Generative Adversarial Network (GAN) \cite{goodfellow2014generative}, there are many identity-preserved masked face restoration methods \cite{din2020novel,ge2020occluded}.
In \cite{din2020novel}, masked face images are first segmented and then impainted with fine facial details while retaining the global coherency of face structure.  Ge \etal~\cite{ge2020occluded} propose identity-preserved inpainting to facilitate occluded face recognition. The core idea is integrating GAN with an optimized pre-trained CNN model which serves as the third player to compete with the generator by enabling the inpainted faces to be close to their identity centers.

Since occlusion recovery methods \cite{din2020novel,ge2020occluded} are more complicated to set up the online evaluation toolkit, we focus on occlusion-robust face feature embedding in this challenge. In \cite{weng2016robust}, a new partial face recognition approach is proposed by using local texture set matching to recognize persons of interest from their partial faces.  In \cite{hariri2021efficient}, a masked-aware face feature embedding is proposed by extracting deep features from the unmasked regions (mostly eyes and forehead regions). In~\cite{montero2021boosting}, masked face augmentation and extra mask-usage classification loss is proposed to train mask robust facial feature embedding. In~\cite{li2021cropping,Wang2020CVPR}, visual attention mechanism is employed to enhance feature learning from non-occluded face regions.

Even though there are some existing explorations for occluded (masked) face recognition, there is yet no publicly available large-scale masked face recognition benchmark due to the sudden outbreak of the epidemic. In this report, we make a significant step further and propose a new comprehensive benchmark for masked face recognition as well as non-masked face recognition. To this end, we have collected a real-world masked test set, children test set, multi-racial test set (\ie African, Caucasian, South Asian and East Asian \cite{xu2020investigating,gong2019jointly,wang2019racial,wang2020mitigating}). We define different sub-tracks with fixed training data, and each sub-track has strict constraints on computational complexity and model size. Therefore, the performance comparison between different models can be fair.

By using the proposed test data, we organized the InsightFace track in Masked Face Recognition Challenge (ICCV 2021). This report presents the details of this track, including the training data, the test set, evaluation protocols, baseline solutions, performance analysis of the top-ranked submissions received as part of the competition, and effective strategies for masked face recognition.
The report of another WebFace260M track is available in \cite{zhu2021mfrwebface}.

\section{Datasets of InsightFace Track}

\subsection{Training Dataset}

As given in Tab. \ref{tab:trainingdatasets}, we employ two existing datasets (\ie MS1M \cite{guo2016ms} and Glint360K \cite{an2020particalfc}) as the training data.

\noindent {\bf MS1M}: The MS1M training dataset is cleaned from the MS-Celeb-1M~\cite{guo2016ms} dataset. All face images are pre-processed to the size of $112\times112$ by the five facial landmarks predicted by RetinaFace~\cite{deng2020retinaface}. Then, a semi-automatic refinement is conducted by employing the pre-trained ArcFace~\cite{deng2019arcface} model and ethnicity-specific annotators~\cite{deng2019lightweight}. Finally, the refined MS1M dataset contains 5.1M images of 93K identities.

\noindent {\bf Glint360K}: The Glint360K training dataset is cleaned from the MS-Celeb-1M~\cite{guo2016ms} and Celeb-500k \cite{cao2018celeb} datasets.
All face images are downloaded from the Internet and pre-processed to the size of $112\times112$ by the five facial landmarks predicted by RetinaFace~\cite{deng2020retinaface}. Then, an automatic refinement is conducted by employing the pre-trained ArcFace~\cite{deng2019arcface} model for intra-class and inter-class cleaning. Finally, the released Glint360K dataset contains 17M images of 360K individuals, which is one of the largest and cleanest training datasets \cite{zhu2021webface260m} in academia.

\begin{table}
\centering
\begin{tabular}{c|c|c}
\hline
Dataset & \# Identities & \# Images\\
\hline
\hline
MS1M & 93K & 5.1M\\
\hline
Glint360K & 360K & 17M  \\
\hline
\end{tabular}
\caption{Statistics of the training data of the masked face recognition challenge (the InsightFace track).}
\label{tab:trainingdatasets}
\end{table}

The training data (\ie MS1M and Glint360K) are fixed to facilitate performance reproduction and fair comparison. Detailed requirements:
\begin{itemize}
\item No external dataset is allowed and no pre-trained model is allowed.
\item All participants must use the predefined training dataset for a particular challenge track. Data augmentation for the facial mask is allowed but the augmentation method needs to be reproducible.
\end{itemize}

\begin{table*}
\begin{center}
\begin{tabular}{c|cc|cc}
\hline
 & \# Identities & \# Images & \# Positive Pairs & \# Negative Pairs \\
\hline\hline
Masked Test Set & 6,964 &20,892 & 13,928 & 96,983,824 \\
\hline\hline
Children Test Set & 14,344 & 157,280 & 1,773,428 & 24,735,067,692 \\
\hline\hline
Multi-racial Test Set &	242,143 &	1,624,305 &	4,689,037 &	2,638,360,419,683\\
\hline
African	  & 43,874  &298,010	& 870,091 &	88,808,791,999 \\
Caucasian &	103,293	& 697,245 &	2,024,609 &	486,147,868,171 \\
South Asian    & 35,086	& 237,080 &	688,259	& 56,206,001,061 \\
East Asian     &	59,890  &	391,970 & 1,106,078	& 153,638,982,852 \\
\hline
\end{tabular}
\caption{Statistics of the test sets of the masked face recognition challenge (the InsightFace track).}
\label{table:teststatistics}
\end{center}
\end{table*}

\begin{figure*}
\centering
\subfigure[{\scriptsize Masked Test Set}]{
\label{fig:s1}
\includegraphics[height=0.35\textwidth]{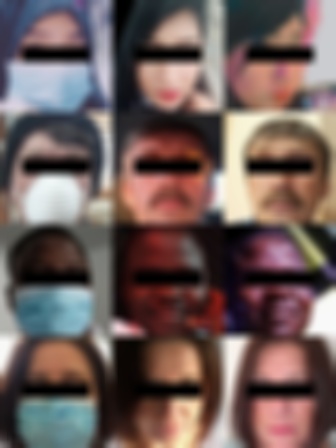}}
\subfigure[{\scriptsize Children Test Set}]{
\label{fig:s2}
\includegraphics[height=0.35\textwidth]{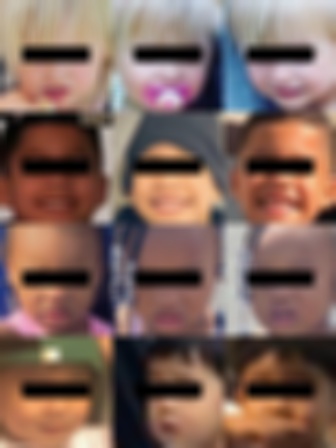}}
\subfigure[{\scriptsize Multi-racial Test Set}]{
\label{fig:s3}
\includegraphics[height=0.35\textwidth]{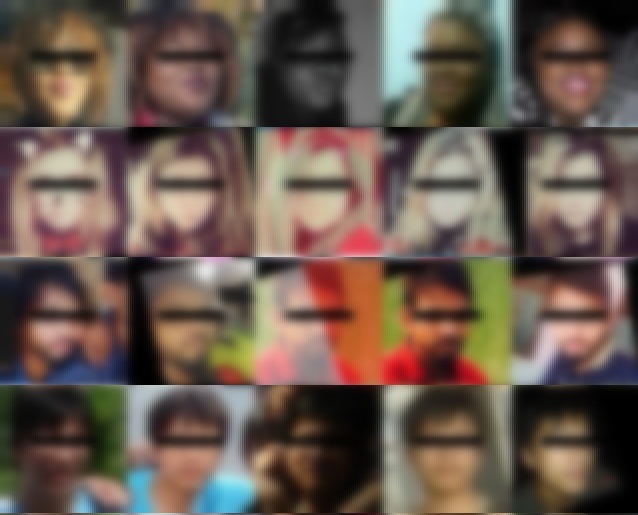}}
\caption{Exemplar blurred face images of Masked Test Set, Children Test Set and Multi-racial Test Set (\ie African, Caucasian, South Asian, East Asian). To ensure data privacy, we intentionally decrease the quality of the exemplar facial images. On our test server, all of the test images are still in high quality.}
\label{fig:sampleimages}
\end{figure*}

\subsection{Test Dataset}

As shown in Tab. \ref{table:teststatistics} and Fig. \ref{fig:sampleimages}, we manually collected the following three test sets for the comprehensive evaluation of different algorithms.
Unlike existing face recognition test sets (\eg LFW \cite{huang2007labeled}, CFP-FP \cite{sengupta2016frontal}, AgeDB \cite{Moschoglou2017AgeDB}, and IJB-C \cite{maze2018iarpa}), our test sets are not collected from celebrities, thus we can naturally avoid the identity-overlapping problem.
The pre-processing step for the test set is the same as that on the training data. All of the faces are normalized into $112\times112$ by using RetinaFace~\cite{deng2020retinaface}. We also employ a semi-automatic method to strictly ensure that (1) most of the test sets are noise-free and (2) there is no identity overlap between our training data and the test set.

\noindent {\bf Masked Test Set}: The masked test set contains 6,964 masked facial images and 13,928 non-masked facial images of 6,964 identities. In total, there are 13,928 positive pairs and 96,983,824 negative pairs for the verification evaluation.

\noindent {\bf Children Test Set}: The children test set contains 157,280 images of 14,344 identities aging between 2 and 16. There are totally 1,773,428 positive pairs and 24,735,067,692 negative pairs for the verification evaluation.

\noindent {\bf Multi-racial Test Set}: Participants will also have their algorithms tested on the multi-racial test set for fairly evaluating the performance on different demographic groups. The multi-racial test set consists of four demographic groups \footnote{Here, we refer to the NIST standard:\\ \url{https://nvlpubs.nist.gov/nistpubs/ir/2019/NIST.IR.8280.pdf}}: African, Caucasian, South Asian and East Asian \cite{xu2020investigating,gong2019jointly,wang2019racial,wang2020mitigating}. In total, there are 1.6M images of 242K identities.

\section{Evaluation Protocols of InsightFace Track}

The test is aimed to determine whether, and to what degree, face recognition performance differs when they process photographs of masked faces, child faces, and individuals from various demographic groups (\eg African, Caucasian, South Asian and East Asian). All pairs between the gallery and probe sets will be used for evaluation. We employ the 1:1 face verification as the evaluation metric.

\noindent {\bf Masked Test Set}: We report True Positive Rate (TPR) @ False Positive Rate (FPR) = 1e-4 given 13,928 positive pairs and 96,983,824 negative pairs.

\noindent {\bf Children Test Set}: We report True Positive Rate (TPR) @ False Positive Rate (FPR) = 1e-4 given 1,773,428 positive pairs and 24,735,067,692 negative pairs.

\noindent {\bf Multi-racial Test Set}: We assess accuracy by demographic groups (\eg African, Caucasian, South Asian and East Asian) and report True Positive Rate (TPR) @ False Positive Rate (FPR) = 1e-6.
The number of positive pairs for each demographic group is of the order of million and the number of negative pairs for each demographic group is of the order of billion.

\noindent {\bf InsightFace Track Ranking Rules}:
To protect data privacy and ensure fairness in the competition, we withhold all images as well as labels of the test data.
Participants can submit their models in the ONNX format to our evaluation server and get their results from the leader-board after the online evaluation (usually several hours). Participants are only allowed to use the training data we provided for a particular challenge track. On the widely used V100 GPU, we set an upper bound of inference time ($<10$ ms/image for the MS1M sub-track and $<20$ ms/image for the Glint360 sub-track) to control the model complexity and the submitted model size should be smaller than 1GB in the format of float32. On our online test server, we employ cosine similarity for the verification test. The feature dimension of the MS1M sub-track should be smaller than $512$ and the feature dimension of the Glint360K sub-track should be smaller than $1024$. All challenge submissions are ordered in terms of weighted TPRs across two test sets (\ie Masked Test Set and Multi-racial Test Set) by the formula of 0.25 * TPR@Masked + 0.75 * TPR@MR-All.

\begin{figure}
\centering
\includegraphics[width=1.0\linewidth]{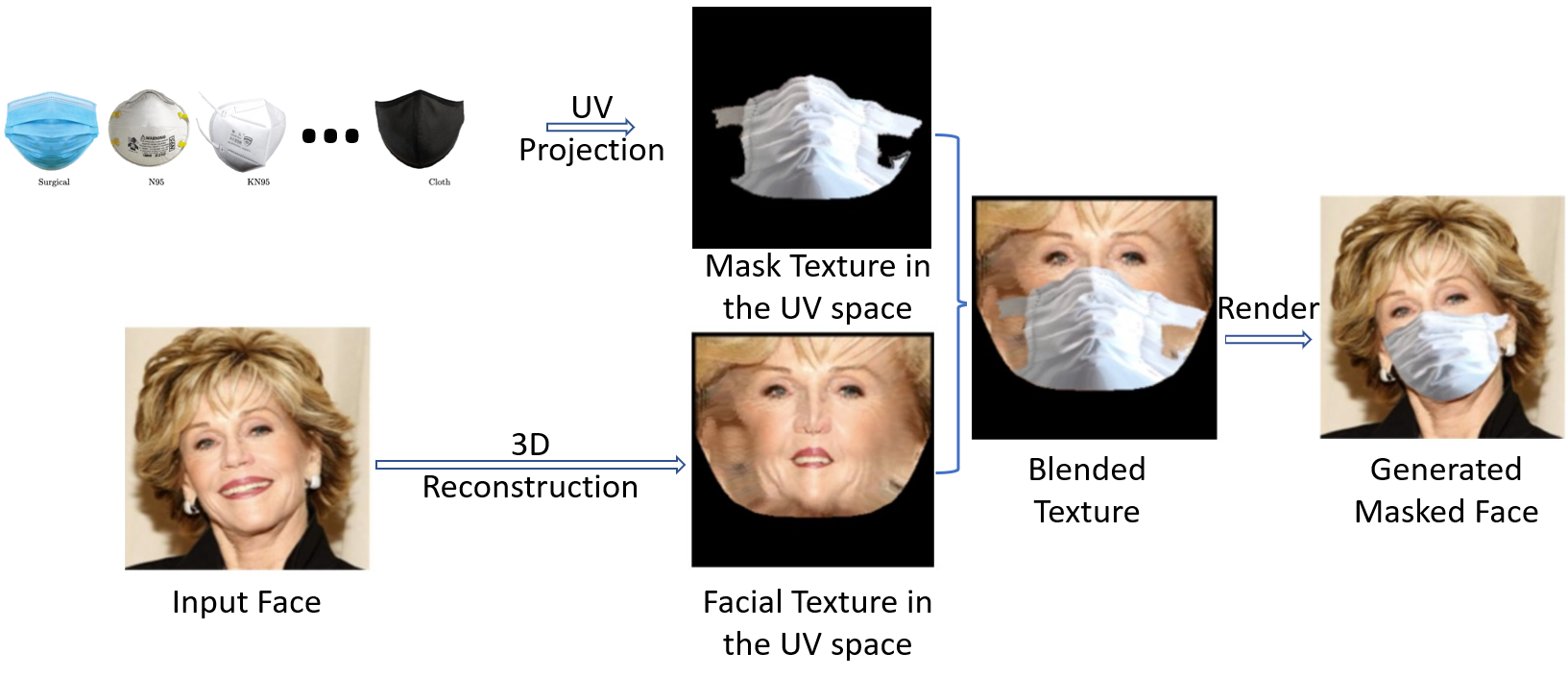}
\caption{Masked face augmentation through texture blending in the UV space. Demo images and the pipeline are from the JDAI-CV toolkit \cite{wang2021facex}.}
\label{fig:maskedfacegeneration}
\end{figure}

\newcommand{\blocka}[2]{\multirow{3}{*}{\(\left[\begin{array}{c}\text{3$\times$3, #1}\\[-.1em] \text{3$\times$3, #1} \end{array}\right]\)$\times$#2}
}
\newcommand{\blockb}[3]{\multirow{3}{*}{\(\left[\begin{array}{c}\text{1$\times$1, #2}\\[-.1em] \text{3$\times$3, #2}\\[-.1em] \text{1$\times$1, #1}\end{array}\right]\)$\times$#3}
}
\renewcommand\arraystretch{1.1}
\setlength{\tabcolsep}{3pt}
\begin{table*}[h]
\begin{center}
\begin{tabular}{c|c|c|c|c|c}
\hline
layer name &  R18 & R34 & R50 & R100 & output size\\
\hline
\hline
 &  & & & & 112$\times$112$\times$3 \\
\hline

stem & 3$\times$3, 64, s=1 & 3$\times$3, 64, s=1 & 3$\times$3, 64, s=1 & 3$\times$3, 64, s=1 & 112$\times$112$\times$64\\

\hline

\multirow{3}{*}{Conv1\_x} & \blocka{64}{2} & \blocka{64}{3}& \blocka{64}{3}& \blocka{64}{3}&  \multirow{3}{*}{56$\times$56$\times$64} \\
  &  &&& & \\
  &  &&& & \\

\hline

\multirow{3}{*}{Conv2\_x} & \blocka{128}{2} & \blocka{128}{4}& \blocka{128}{4}& \blocka{128}{13}&  \multirow{3}{*}{28$\times$28$\times$128} \\
  &  &&& & \\
  &  &&& & \\

\hline
\multirow{3}{*}{Conv3\_x}   & \blocka{256}{2} & \blocka{256}{6}& \blocka{256}{14}& \blocka{256}{30}& \multirow{3}{*}{14$\times$14$\times$256}\\
  & &&&  & \\
  & &&&  & \\

\hline
\multirow{3}{*}{Conv4\_x} & \blocka{512}{2} & \blocka{512}{3}& \blocka{512}{3}& \blocka{512}{3} & \multirow{3}{*}{7$\times$7$\times$512}  \\
  & &&&  & \\
  & &&&   & \\
\hline
  FC & &&& & 1$\times$1$\times$512 \\
 \hline\hline
Flops (G) & 2.62 & 4.48 & 6.33 & 12.12 & \\
\# Params (M)& 24.03 & 34.14 & 43.59 & 65.16 &  \\
\hline
\end{tabular}
\end{center}
\caption{The network configuration, computation complexity and model size of baseline models. Convolutional building blocks are shown in brackets with the numbers of blocks stacked. Down-sampling is performed by the second conv in conv1\_1, conv2\_1, conv3\_1, and conv4\_1 with a stride of 2.}
\label{tab:backboneconfig}
\end{table*}

\begin{table*}
\begin{center}
\begin{tabular}{ccc|c|c|c|cc}
\hline
Data & Backbone & Loss &  LFW & CFP-FP & AgeDB & IJB-C (1e-4) & IJB-C (1e-5)\\
\hline
MS1M & R18  & ArcFace & 99.77 & 97.73 & 97.77 & 94.66 & 92.07\\
MS1M & R34  & ArcFace & 99.80 & 98.67 & 98.10 & 95.90 & 94.10\\
MS1M & R50  & ArcFace & 99.83 & 98.96 & 98.35 & 96.46 &	94.79\\		
MS1M & R100 & ArcFace & 99.85 &	99.06 & 98.48 & 96.81 &	95.31\\			
\hline\hline
Glint360K & R18 & ArcFace  & 99.77 & 97.73 & 97.72 & 95.33	& 93.16\\
Glint360K & R34 & ArcFace  & 99.82 & 98.78 & 98.33 & 96.56	& 95.16\\		
Glint360K & R50 & ArcFace  & 99.83 & 99.20 & 98.38 & 96.97  & 95.61\\				
Glint360K & R100 & ArcFace & 99.82 & 99.29 & 98.48 & 97.32  & 95.88\\			
\hline
\end{tabular}
\caption{Baseline performance on the public test benchmarks (\eg LFW, CFP-FP, AgeDB and IJB-C).}
\label{table:baselineperformanceonpublic}
\end{center}
\end{table*}

\section{Baseline Solutions of InsightFace Track}

Training details of baseline models are released before the challenge to facilitate participation. We re-implement a simple online masked face augmentation function \cite{wang2021facex}, customize the ResNet~\cite{he2016deep} for the baseline models and employ ArcFace~\cite{deng2019arcface} as our loss function, which is one of the top-performing methods for deep face recognition.

\subsection{Masked Face Augmentation}

As shown in Fig.~\ref{fig:maskedfacegeneration}, we follow the JDAI-CV toolkit \cite{wang2021facex} \footnote{\url{https://github.com/JDAI-CV/FaceX-Zoo/tree/main/addition_module/face_mask_adding/FMA-3D}} to implement our online masked face generation function \footnote{\url{https://github.com/deepinsight/insightface/tree/master/recognition/\_tools\_}}. After 3D face reconstruction \cite{feng2018joint} on the input 2D face image, we obtain the UV texture map, the face geometry and the camera pose. Then, we randomly select one facial mask from the collected mask dataset and project it into the UV space. Based on a simple texture blending, we can easily get the masked facial UV texture. Finally, we combine the masked facial UV texture and the face geometry, and render the masked face into a 2D face image.

\subsection{Implementation Details}

During training, we follow ArcFace~\cite{deng2019arcface} to set the feature scale to $64$ and choose the angular margin at $0.5$.
As shown in Tab.~\ref{tab:backboneconfig}, we customize the ResNet~\cite{he2016deep} as our baseline models (\ie R18, R34, R50 and R100). More specifically, we only employ the basic residual block instead of the bottleneck residual block following ArcFace~\cite{deng2019arcface}. The baseline models are implemented by PyTorch with parallel acceleration on both features and centres\footnote{\url{https://github.com/deepinsight/insightface/tree/master/recognition/arcface\_torch}}. We set batch size as $1,024$ and train models on eight NVIDIA V100(32GB) GPUs. The learning rate starts with $0.1$, drops by $0.1$ at 10, 16, 22 epochs, and the whole training procedure finishes at 24 epochs. We set the momentum to $0.9$ and the weight decay to $5e-4$. During testing, we only keep the feature embedding network without the fully connected layer and extract the $512$-$D$ features for each normalized face crop. We use the cosine similarity metric for each feature pair.

\subsection{Baseline Performance}
As shown in Tab.~\ref{table:baselineperformanceonpublic}, we first test our baseline models on public benchmarks, including LFW \cite{huang2007labeled}, CFP-FP \cite{sengupta2016frontal}, AgeDB \cite{Moschoglou2017AgeDB}, and IJB-C \cite{maze2018iarpa}. By increasing the computation complexity from R18 to R100, the performance rises steadily across all test sets. After changing the training data from MS1M to Glint360K, TPR@FPR=1e-4 on IJB-C significantly increases from $96.81\%$ to $97.32\%$ for R100. On CFP-FP, R100 trained on Glint360K outperforms the counterpart model trained on MS1M by $0.23\%$, indicating that frontal-to-profile face verification can benefit from more training data. By contrast, the verification accuracy on LFW and AgeDB is almost the same, which indicates that LFW and AgeDB are saturated to distinguish high performing models.

In Tab.~\ref{table:baselineperformance}, we report the performance on the challenge benchmarks.
As we can see from these results, the verification accuracy benefits from more training data (from MS1M to Glint360K) and heavier backbone structures (from R18 to R100) across all testing scenarios (\ie masked, children and multi-racial test sets). Compared to the performance gaps on public test datasets (\ie LFW \cite{huang2007labeled}, CFP-FP \cite{sengupta2016frontal}, AgeDB \cite{Moschoglou2017AgeDB}, and IJB-C \cite{maze2018iarpa}), the performance gaps on the proposed masked test set, children test set and multi-racial test set are more obvious. In addition, we also conduct experiments with masked face augmentation. When $10\%$ of training data wear facial masks during training, the verification accuracy on the masked test set significantly increases from $69.091\%$ to $77.325\%$ by using MS1M, and increases from $75.567\%$ to $83.710\%$ by using Glint360K. However, masked face augmentation is slightly harmful for non-masked face verification, as the TPR on the MR-All dataset drops by $0.484\%$ for the MS1M sub-track and decreases by $0.644\%$ for the Glint360K sub-track. We leave the balance of masked face augmentation for the challenge participants.

\begin{table*}
\begin{center}
\resizebox{\linewidth}{!}{
\begin{tabular}{ccc|c|c|ccccc|cc}
\hline
Data & Backbone & Loss &  Mask	& Children  &	African &	Caucasian	& South Asian &	East Asian &	MR-All & Size(MB) &	Time (ms)\\
\hline
MS1M & R18  & ArcFace & 47.853 & 	41.047 & 	62.613 &	75.125 &	70.213 &	43.859 &	68.326 &	91.658	& 1.856\\
MS1M & R34  & ArcFace & 58.723 &    55.834 &	71.644 &	83.291 &	80.084 &	53.712 &	77.365 &	130.245 & 3.054\\
MS1M & R50  & ArcFace & 63.850 &	60.457 &	75.488 &	86.115 &	84.305 &	57.352 &	80.533 &	166.305	& 4.262\\
MS1M & R100 & ArcFace & 69.091 &	66.864 &	81.083 &	89.040 &	88.082 &	62.193 &	84.312 &	248.590	& 7.031\\
\hline
MS1M+MA-0.1 & R100 & ArcFace & 77.325 &	67.053 &	80.247 &	88.706 &	87.583 &	61.410 &	83.828 &	248.590	& 7.032\\
\hline\hline
Glint360K & R18 & ArcFace  & 53.317 &	48.113 &	68.230 &	80.575 &	75.852 &	47.831 &	72.074 &	91.658  &	2.013 \\
Glint360K & R34 & ArcFace  & 65.106 &	65.454 &	79.907 &	88.620 &	86.815 &	60.604 &	83.015 &	130.245 &	3.044 \\
Glint360K & R50 & ArcFace  & 70.233 &	69.952 &	85.272 &	91.617 &	90.541 &	66.813 &	87.077 &	166.305 &	4.340 \\
Glint360K & R100 & ArcFace & 75.567 &	75.202 &	89.488 &	94.285 &	93.434 &	72.528 &	90.659  &	248.590	&   7.038 \\
\hline
Glint360K+MA-0.1 & R100 & ArcFace & 83.710 &	75.894 &	88.919 &	94.038 &	92.882 &	71.137 &	90.015 &	248.590	& 7.036\\
\hline
\end{tabular}
}
\caption{The baseline performance of the masked face recognition challenge (the InsightFace track). ``MR-All'' denotes the verification accuracy on all multi-racial images. Inference time is evaluated on Tesla V100 GPU using onnxruntime-gpu==1.6. ``MA-0.1'' means masked face augmentation with a specific probability of $10\%$.}
\label{table:baselineperformance}
\end{center}
\end{table*}

\begin{table*}
\begin{center}
\resizebox{\linewidth}{!}{
\begin{tabular}{c|c|c|c|ccccc|ccc}
\hline
Rank & Participant &  Mask	& Children  &	African &	Caucasian	& South Asian &	East Asian &	MR-All & Size(MB) &	Time (ms) & Feat Dim\\
\hline
1 & agir & 84.169 & 75.003 & 88.322 & 93.396 & 93.349 & 72.623 & 90.452 & 317.665 & 9.764 & 512 \\
2 & Rhapsody & 83.831 & 64.152 & 86.516 & 93.459 & 92.461 & 72.616 & 90.098 & 327.618 & 9.083 & 512 \\
3 & paradox & 84.183 & 75.105 & 88.436 & 93.374 & 92.398 & 71.127 & 89.710 & 357.488 & 9.520 & 512 \\
4 & mayidong & 84.312 & 73.936 & 86.258 & 92.227 & 91.244 & 70.042 & 88.897 & 295.954 & 9.762 & 512 \\
5 & jerrysunnn & 82.201 & 57.467 & 85.395 & 92.124 & 91.270 & 71.501 & 89.252 & 327.624 & 9.036 & 512 \\
6 & mind\_ft & 84.528 & 68.303 & 86.820 & 92.251 & 88.326 & 68.595 & 88.355 & 250.145 & 9.318 & 512 \\
7 & upupup & 82.352 & 53.794 & 85.069 & 92.061 & 91.044 & 71.159 & 89.000 & 327.618 & 9.071 & 512 \\
8 & hammer\_hk & 81.706 & 58.097 & 84.853 & 91.917 & 91.163 & 70.783 & 88.894 & 327.618 & 9.078 & 512 \\
9 & unitykd0701 & 83.522 & 71.915 & 84.158 & 91.172 & 89.093 & 68.684 & 87.239 & 322.265 & 9.656 & 512 \\
10 & kisstea & 83.831 & 71.050 & 83.828 & 90.866 & 90.054 & 67.108 & 87.046 & 288.849 & 9.079 & 512 \\
11 & Hello & 79.308 & 67.126 & 86.012 & 92.168 & 92.603 & 68.694 & 88.529 & 250.145 & 9.253 & 512 \\
12 & JulieXU & 82.209 & 70.465 & 83.823 & 90.734 & 89.889 & 68.194 & 87.236 & 235.505 & 7.909 & 512 \\
13 & hjgw & 82.115 & 70.463 & 83.813 & 90.689 & 89.956 & 68.039 & 87.155 & 235.505 & 7.892 & 512 \\
14 & xuyang1 & 76.163 & 77.104 & 87.962 & 93.256 & 92.580 & 68.774 & 89.080 & 302.869 & 9.654 & 512 \\
15 & webill & 78.123 & 72.833 & 85.942 & 92.099 & 91.151 & 69.273 & 88.333 & 253.756 & 9.688 & 512 \\
\hline
\end{tabular}
}
\caption{Top-15 submissions of the MS1M sub-track by 16 August 2021.}
\label{table:leaderboardms1m}
\end{center}
\end{table*}

\begin{table*}
\begin{center}
\resizebox{\linewidth}{!}{
\begin{tabular}{c|c|c|c|ccccc|ccc}
\hline
Rank & Participant &  Mask	& Children  &	African &	Caucasian	& South Asian &	East Asian &	MR-All & Size(MB) &	Time (ms) & Feat Dim\\
\hline
1 & jerrysunnn & 88.972 & 86.628 & 93.064 & 96.278 & 95.578 & 77.969 & 93.512 & 728.879 & 18.981 & 512 \\
2 & mayidong & 86.933 & 84.545 & 93.043 & 96.529 & 95.361 & 77.554 & 93.358 & 541.023 & 17.837 & 1024 \\
3 & derron & 87.636 & 84.679 & 92.289 & 95.564 & 95.109 & 77.512 & 92.815 & 566.625 & 19.153 & 1024 \\
4 & mind\_ft & 86.962 & 81.321 & 92.566 & 96.100 & 95.417 & 76.024 & 92.705 & 456.505 & 17.801 & 512 \\
5 & DongWang & 83.487 & 82.619 & 92.182 & 95.655 & 94.791 & 76.392 & 92.743 & 564.831 & 17.101 & 1024 \\
6 & helloface & 87.881 & 82.423 & 90.337 & 94.382 & 93.272 & 73.818 & 91.252 & 248.590 & 7.025 & 512 \\
7 & didujustfart & 89.130 & 83.165 & 90.373 & 95.181 & 93.557 & 72.638 & 90.776 & 453.391 & 13.584 & 512 \\
8 & yossi\_avram & 81.469 & 82.848 & 92.974 & 96.162 & 95.739 & 76.584 & 93.005 & 248.583 & 7.472 & 512 \\
9 & tinytan & 84.657 & 81.438 & 91.139 & 95.183 & 94.238 & 75.369 & 91.823 & 497.807 & 14.875 & 1024 \\
10 & deepcam & 84.391 & 82.781 & 90.871 & 94.573 & 94.176 & 75.821 & 91.834 & 284.102 & 11.276 & 512 \\
11 & sgglink & 84.966 & 78.323 & 90.067 & 94.738 & 93.251 & 73.284 & 91.162 & 605.258 & 19.071 & 512 \\
12 & suanying & 77.642 & 82.362 & 92.874 & 96.305 & 95.484 & 77.582 & 93.477 & 456.505 & 17.813 & 512 \\
13 & dingweichao & 82.223 & 79.652 & 90.042 & 94.772 & 93.438 & 73.479 & 91.263 & 453.391 & 13.559 & 512 \\
14 & HYL\_Dave & 78.116 & 79.770 & 91.670 & 95.575 & 94.690 & 74.872 & 92.098 & 453.391 & 13.602 & 512 \\
15 & EvilGeniusFeng & 76.249 & 80.140 & 91.042 & 94.860 & 94.300 & 75.468 & 92.011 & 250.390 & 9.363 & 512 \\
\hline
\end{tabular}
}
\caption{Top-15 submissions of the Glint360K sub-track by 16 August 2021.}
\label{table:leaderboardglint360k}
\end{center}
\end{table*}

\section{Leader-board Results of InsightFace Track}

The masked face recognition competition (InsightFace track) is conducted as part of the \textit{Masked Face Recognition Challenge \& Workshop}\footnote{\url{https://ibug.doc.ic.ac.uk/resources/masked-face-recognition-challenge-workshop-iccv-21/}}, at the International Conference on Computer Vision 2021 (ICCV 2021). Participants can freely select different sub-tracks to develop a face feature embedding model, which is automatically evaluated on our test server based on the above-mentioned protocols. The competition has been opened worldwide, to both industry and academic institutions. By 16th August 2021, the InsightFace track has received hundred of registrations from across the world. More specifically, the competition has received 123 valid submissions for the MS1M sub-track and 69 valid submissions for the Glint360K sub-track. Here, multi-submissions for one sub-track from the same participant is only counted once.

As we postpone the leader-board submission to 1st October 2021, we can not collect the final top-ranked solutions before the camera-ready deadline. After the competition, we will close the test server and select the valid top-3 solutions for each track. We will collect the training code from these top-ranked participants and re-train the models to confirm whether the performance of each submission is valid or not. We will update the challenge report through arxiv with detailed team information and detailed top-ranked solutions.

By 16th August 2021, we have found the best model of the MS1M sub-track has achieved $84.169\%$ on the masked test set, and $90.452\%$ on the MR-All test set.
As given in Tab.~\ref{table:leaderboardms1m}, we list the top-15 submissions from the leader-board. Comparing with the baseline models in Tab.~\ref{table:baselineperformance},
there are around $7\%$ absolute improvements on the masked test set and the MR-All test set. For the Glint360K sub-track, the best model has achieved $ 88.972\%$ on the masked test set, and $93.512\%$ on the MR-All test set as shown in Tab.~\ref{table:leaderboardglint360k}. Comparing with the baseline models in Tab.~\ref{table:baselineperformance},
there is around $6\%$ absolute improvement on the masked test set and about $3.5\%$ absolute improvement on the MR-All test set. Therefore, there is huge space for the training optimization to improve masked face recognition without the accuracy drop on the non-masked face recognition.

\section{Ethical Considerations}

Face recognition has been a controversial topic recently. There have been questions over ethical concerns about invasion of privacy, alongside how well face recognition systems recognize darker shades of skin (known as the bias problem).
In the InsighFace track of this challenge, we employ existing academic data as the training datasets. Most of the identities inside the training data are well-known celebrities~\cite{guo2016ms}. The pre-processed training data are compressed into the binary record and only released to relevant researchers to facilitate the reproducible training. Our private test data will not be released to the public to avoid the data privacy problem and ensure fairness for all participants.
For the bias concern, we follow the most authoritative evaluation set up by NIST-FRVT \footnote{\url{https://nvlpubs.nist.gov/nistpubs/ir/2019/NIST.IR.8280.pdf}}. We wish to promote fairness among deep face recognition and thus set up the multi-racial verification benchmark.

\section{Conclusions and Future Works}

In this InsightFace track report, we introduce our new benchmark for the evaluation of masked face recognition as well as non-masked face recognition. Based on our baseline solutions,
we confirm the effectiveness of the naive masked face augmentation. As the challenge is still under-going, we will keep on updating the top-ranked solutions as well as this report on arxiv.

Besides the InsightFace track, there is also a parallel WebFace260M track \footnote{\url{https://www.face-benchmark.org/challenge.html}} in the Masked Face Recognition challenge. The WebFace260M track is organized by Zheng Zhu, Guan Huang, Jiankang Deng, Yun Ye, Junjie Huang, Xinze Chen, Jiagang Zhu, Tian Yang, Jia Guo, Jiwen Lu, Dalong Du and Jie Zhou. Detailes can be found in the arxiv report~\cite{zhu2021mfrwebface}, which will be also updated in the future.

{\small
\bibliographystyle{ieee_fullname}
\bibliography{egbib}
}
\end{document}